\documentclass{article}
\usepackage{graphicx}
\usepackage{amsfonts}
\newcommand{\R}{\mathbb{R}}

\begin{document}
\title{The Structure of Narrative: the Case of Film Scripts}

\author{Fionn Murtagh (1), Adam Ganz (2) and 
Stewart McKie (2) \\
(1) Department of Computer Science \\
(2) Department of Media Arts \\
Royal Holloway, University of London, Egham TW20 0EX, England \\
Contact author: fmurtagh at acm dot org}

\maketitle

\begin{abstract} 
We analyze the style and structure of story narrative using the case 
of film scripts.  The practical importance of this is noted, especially 
the need to have support tools for television movie writing.  
We use the Casablanca film script, and scripts from six episodes
of CSI (Crime Scene Investigation).  For analysis of style and 
structure, we quantify various central perspectives discussed in McKee's 
book, {\em Story: Substance, Structure, Style, and the Principles
of Screenwriting}.  Film 
scripts offer a useful point of departure for exploration of the 
analysis of more general narratives.  Our methodology, using 
Correspondence Analysis, and hierarchical clustering, is innovative 
in a range of areas that we discuss.  In particular this work is 
groundbreaking in taking the qualitative analysis of McKee and 
grounding this analysis in a quantitative and algorithmic framework.
\end{abstract}

\noindent 
{\bf Kewwords:}
data mining, data analysis, factor analysis, correspondence 
analysis, semantic space,
Euclidean display, hierarchical clustering, narrative, story,
film script

\section{Introduction}
\label{intro}

As a framework for analysis of narrative in many areas of application, 
film scripts have a great deal to offer.  We present quite a number of
innovations in this work: quantifying and automating a range of 
qualitative ways of addressing pattern recognition in narrative; use 
of metric embedding using Correspondence Analysis, and ultrametric 
embedding through hierarchical 
clustering of a data sequence, in order to capture semantics in the 
data; and verifying experimentally the well foundedness of much that 
McKee \cite{mckee} describes in qualitative terms.  

Other than the data mining, two distinct levels of user are at issue here.  
Firstly and foremostly, we have the scriptwriter or screenwriter in mind.  
See \cite{mckee} for a content-based description of the scriptwriting process.
Secondly and more indirectly we have the movie viewer in mind.  The 
feasibility of using statistical learning methods in order to map a 
characterization of film scripts (essentially using 22 characteristics)
onto box office profitability was pursued by \cite{eliashberg}.  The 
importance of such machine learning of what constitutes a good quality
and/or potentially profitable film script has been the basis for 
(successful) commercial initiatives, as described by \cite{gladwell}.  
The business side of the movie business is elaborated on in some depth 
in terms of avenues to be explored in the future, in \cite{eliashberg2}.

A film script is semi-structured in that it is subdivided into scenes 
and sometimes other structural units.  Furthermore there is metadata 
provided related to location (internal, external; particular or general 
location name); characters; day, night.  There is also dialog and 
description of action in free text.  
 
There are literally thousands of film scripts, for all genres, available 
and openly accessible (e.g.\ IMSDb, The Internet Movie Script Database,
www.imsdb.com).  While offering just one data modality, viz.\ text, 
there is close linkage to other modalities, visual, speech and often music.  

An area of application of our work that is of particular importance 
is to movies for television.
In contrast to a  cinema movie, in television a serial dramatization 
is formulaic in its set of characters, their actions, location, 
and in content generally.
In form it is even more formulaic, in length, initial scenes, positioning 
of advertisement breakpoints, and in other aspects of format.  While 
screenwriting and subsequent aspects of cinema movie creation and 
development have been much studied (e.g.\ \cite{mckee}), film 
writing for  television has been less so.  The latter is often 
team-based, and much more ``managed'' on account of its linkages 
with other dramatizations in the same series, or in closely related 
series.  These properties of television serial dramatization  lead, 
perhaps even more than for cinema movie,  
to a need for tools to quantitatively support script writing.  

Film scripts constitute an outstanding template or model for other 
domains.  
One longer term goal of our work is for our tools to provide a platform
for introducing interactivity into a movie.  
Interactivity with a film or video includes the following possible
uses: (i) it provides an approach to developing interactive games;
(ii) it allows for use in interactive and even immersive training and
learning environments; and (iii) support is also feasible for use in
the entertainment area, e.g. interactive television. Leveraging the
existing corpus and ``fanbase'' of hit scripts to create new interactive
versions could create a whole new sector of movie script based interactive
games or filmed variant-sequels to famous movies that take the familiar
in a different direction.
The convergence of story-telling and narrative, on the one hand, and
games, on the other, is explored in \cite{glassner}.  There and in 
\cite{riedl}, story graphs and narrative trees are used as a way to 
open up to the user 
the range of branching possibilities available in the story.  
A further longer term goal lies in the area of business 
analysis, especially in distributed, video-conferencing and other, 
settings.  The narrative structure, similar to a film script, gives 
the analyst the framework for generating stories.  Such narrative 
structures can be used for analysis of meetings, projects, dialogs,
and any interactive sessions in real or virtual meetings.  

Our work is hugely topical. A television serial episode costs 
between \$2 to \$3 million for one hour of television.  New series 
are scripted and run on television channels and only then is 
viewer reaction used to determine their viability.  This points 
to major potential savings. Understanding and supporting 
scriptwriting is crucial.  An additional motivation is that the 
entertainment and cultural market is moving towards greater 
interactivity, and our work is of great potential here too.  

The structure of the paper is as follows.  In section \ref{delta},
we provide essential background on the analysis approach taken.  
In section \ref{justification}, we discuss the plausibility of 
taking text as a proxy for, or a practical and useful 
expression of, underlying content that we refer to as the story.  
In section \ref{methodology}, we describe briefly the data
 analysis algorithms that we use, providing citations to more detailed
background reading.  

In section \ref{casablanca} we study a number of aspects of 
the Casablanca movie script.  In section \ref{csi}, we turn attention 
to the television series, CSI (Crime Scene Investigation, Las Vegas).   

Among important aspects of this work are the following.  Firstly,
we recast a number of issues discussed by McKee \cite{mckee} in a 
quantitative and algorithmic form.  Secondly, we employ Correspondence
Analysis as a versatile data analysis framework.  Since in 
Correspondence Analysis, each scene is expressed as an average of
attributes used to characterize the scenes, and since each 
attribute is expressed as an average of the scenes they 
characterize, we have in Correspondence Analysis a way to 
represent and study semantics.  Thirdly, our 
clustering of film script units (e.g., scenes) is innovative in a
few ways, including respecting the sequence of film script units, 
and taking as input the ``direction'' of the film script content 
rather than having  a more static framework for the input (which we
found empirically to work less well in that it was far less 
discriminatory).  This type of clustering captures the semantics
of change.  If scenes are very dissimilar they will be agglomerated
late in the sequence of agglomerations.  

\section{Analysis of Change in Content Over Time} 
\label{delta}

Analysis of a film script has everything to do with change in 
the course of the story, as we will now discuss,
following McKee \cite{mckee}.  ``The finest writing ... arcs or changes ...
over the course of the telling'' (p. 104, \cite{mckee}).

A typical film has 40 to 60 ``story events'' or {\em scenes}. 
A scene is a story in miniature
and must have activity or change. 
A scene will typically (with, of necessity, large variability)
translate into 2 to 3 minutes of film. 
In our work we have examples of very short (one or two sentence)
dialog or action descriptions in scenes.  Compared to longer 
scenes, such short scenes can be equally revealing and significant.
Units of action or behavior within a scene are termed {\em beats}
by 
McKee.  We will examine beats below in  a case study using the
Casablanca script.  Ideally every scene is a turning point, in
character behavior, or in changing the ``values'' involved from
positive to negative or vice versa.

 Moving upwards now in scale of units of film script, a {\em sequence}
is a series of typically 2 to 5  scenes.  A sequence expresses
significant albeit moderate change.  
An {\em act} is a set of sequences expressing major impact.
Acts are the ``macro-structure of story'' (p. 217, \cite{mckee}).
It is possible to have a one-act play or story, although this would be
quite unusual.  

The overall set of scenes (or sequences, or acts) is termed the
{\em plot}.  
A climax in a sequence or act or plot  is essential. It is
final and irreversible (pp.\ 41--42, \cite{mckee}).  It is
possible to distinguish between an open and a closed ending,
respectively bringing finality versus leaving some issues hanging,
and this does not alter the all-important climax. 
A sequence climax is of moderate importance, an act climax is of
greater importance, and the plot climax is crucial.  
In this scheme of things, each scene is a turning point
of limited, moderate or great importance in the context of the
overall plot or story.

The degree of change is the difference between a scene and the
scene that climaxes a sequence, or the scene that climaxes an act,
or the overall and global climax of the plot.
In the plot, there is no room for a missing scene, or
a superfluous scene, or a misordering of scenes. 

Rhythm and tempo are related to scene length.  The former, rhythm, 
should be
very variable in order to keep the viewer's (reader's) attention, 
allowing for a soft upper limit on the time-span of attention.  
Tempo can exemplify climax in two ways: firstly,  
through increasingly shortened 
scenes (or other units) leading up to a climax; and, secondly, 
through the climax 
being of clearly greater length than the preceding scene (in order
to allow for sufficient time to elaborate and offload its content). 
A decrease in tempo, through a sudden increase in  scene length, 
can be indicative of 
d\'enouement and climax.  Increasing tempo, manifested through 
successively decreasing scene length, expresses a build-up of tension.  

Our discussion up to now has been about story.  We turn to types
 of structure.
McKee \cite{mckee} categorizes design into classical; minimalist or
miniplot; and anti-structure or antiplot.   In this categorization
scheme for film, the minimalist design
dovetails very well with television film and drama, with e.g.\ open ending,
unreconciled internal conflict, and passive and/or multiple protagonists.  
Below we will
explore case studies based on episodes of CSI (Crime Scene
Investigation).

\section{Justification for Textual Analysis of Film Script}
\label{justification}

In this section we address the issue of plausibility of {\em 
appreciable}
analysis of content based on what are ultimately the statistical
frequencies of co-occurrence of words.  

Words are a means or a medium for getting at the substance and
energy of a story (p. 179, \cite{mckee}).
Ultimately sets of phrases express such underlying issues
(the ``subtext'', as expressed by McKee, a term we avoid due to
possible confusion with subsets of text) as
conflict or emotional connotation (p. 258).
We have already noted that change and evolution is inherent to a
plot.  Human emotion is based on particular transitions. 
So this establishes well the possibility that words and phrases
are not taken literally 
but instead can appropriately capture and represent such transition.
Text, says McKee, is the ``sensory surface'' of a work of art
(counterposing it to the subtext, or underlying emotion or
perception).

Simple words can express complex underlying reality.
Aristotle, for example, used words in common usage to express
technically loaded  concepts (\cite{ref08888}, p. 169), and Freud did
also.

Best practice in film script writing includes the following.
Present tense dominates all (\cite{mckee}, p. 395):
``The ontology of the screen is {\em an absolute present tense in constant
vivid movement}'' (emphasis in original). 
Clipped diction is needed.
Generic nouns are avoided in favor of specific terms, and similarly
adjectives and adverbs are to be avoided.
The verb ``to be'' is to be avoided because: ``Onscreen nothing is in a
state of being; story life is an unending flux of change, of
becoming'' (p. 396).  Simile and metaphor are out, as is any
explicit positioning of context on behalf of the reader of the script
or viewer of the film.  

\section{Methodology: Euclidean Embedding through Correspondence
Analysis, and Clustering the Succession of Film Script Scenes}
\label{methodology}

We have already noted (section \ref{intro}) some novel aspects
of our methodology.  We begin with the display of data (e.g., 
scenes and/or words) where visualization of relationships is 
greatly facilitated by having a Euclidean embedding.  
We show how Correspondence Analysis furnishes such 
a metric space embedding of
the information present in the film script text, and furthermore how this
facilitates an ultrametric (i.e.\ hierarchical) embedding that takes
account of the temporal, semantic dynamic of the film
script narrative.

\subsection{A Note on Correspondence Analysis}
\label{sect21}

Correspondence Analysis \cite{ref08888} takes input data 
in the form of frequencies of
occurrence, or counts, and other forms of data, and produces 
such a Euclidean embedding.  
The Appendix provides a short introduction to 
Correspondence Analysis and hierarchical clustering.  

We start with a cross-tabulation of a set of observations
and a set of attributes.  This starting point is 
an array of counts of presence versus absence, or frequency of
occurrence.  From this input data, 
we can embed the observations and attributes in a
Euclidean space.  This factor space is mathematically optimal in a certain
sense (using the least squares criterion, which is also Huyghens' principle of
decomposition of inertia).  Furthermore a
Euclidean space allows for easy visualization that would be more awkward
to arrange otherwise.  

A third reason for the particular embedding used for the observations and
attributes in the Euclidean factor space is that weighting of observations
and attributes is handled naturally in this framework.
The issue of weighting has to be addressed somehow, with one 
option being to treat the
set (of observations or of attributes) as identically weighted, and hence of
equal a priori importance.  Very often either observations or attributes
follow a  power law: examples of such power laws include Zipf's law (in
natural language texts, the word frequency is inversely
proportional to its rank), the Pareto distribution in economics, and
many others \cite{newman}.  Correspondence Analysis handles weighting of
observations and attributes at the ``core'' of its algorithm \cite{ref08888}.

Application of the power law property to
social networks has included the network of movie actors \cite{newman}.
As counterposed to such work, 
our interest is in analyzing a time series of data.
The succession of scenes in a movie, or acts in a play, exemplify this.

\subsection{Input Data Used}

We take all words into account in the semi-structured texts that 
are provided by film (or television program) scripts.  
Punctuation is ignored.  Upper case is converted to lower case for 
our purposes.  Words must be at least two characters in length.
Any numerical figure, or term beginning with a numerical, is 
ignored.  We have already noted above (section \ref{justification}; 
see also the case studies and discussion in \cite{ref08888})
why the use of all words in this way is, in principle, feasible and
justified.  

Occurrences
or presence/absence data therefore are the point of departure.  
Each scene is cross-tabulated by the set of all words so that, in 
this cross-tabulation table, at the intersection of scene $i$ and word $j$ 
we have a presence (1) or absence (0) value.  
To employ notions of change or proximity between scenes, we need
this data to be appropriately represented in a numerically 
well-defined semantic space.  This is provided by mapping the
frequencies of occurrence data into a Euclidean space, using
Correspondence Analysis.  

\subsection{Hierarchical Agglomerative Clustering Respecting 
Sequence}
\label{cclink}

Unusually in this work, we use hierarchical clustering 
taking a {\em sequence} of film script units (e.g., scenes) into 
account.  Our motivation is the following: large ultrametric 
or tree distances derived from the hierarchy have a ready 
interpretation in terms of {\em change}.  
A brief introduction is provided in the Appendix.  A 
short discussion follows.  

Hierarchical clustering is carried out through a sequence 
of agglomerations of successive scenes or of clusters 
(or temporal segments or intervals) of successive scenes.  
Respect for the sequentiality of the scenes is ensured through
the requirement that agglomerands must be adjacent.  In addition 
to this, the least dissimilar scenes are checked, with the criterion
being: merge two clusters when the {\em greatest} dissimilarity between
 cluster members (scenes) is minimal.  The least
dissimilar pair of scenes, considering two potential agglomerands,
is what we use as our agglomeration criterion.  This is the 
contiguity-constrained complete link hierarchical clustering
method.  Apart from heuristic reasons for favoring it, it has some
further properties that will now be discussed briefly.  

The sequence-related adjacency requirement must take into consideration
whether an agglomerative
clustering method would give rise to an inversion, i.e., a later 
agglomeration in the sequence of agglomerations would have an associated
criterion value that is less than the previous criterion value.  It is 
not hard to appreciate that our desire to have gradations of distance 
represented by the dendrogram would be negated, and severely so, 
by such absence of the monotonicity of criterion value, which amounts to a 
contradiction in interpretation of the dendrogram.

It is shown in \cite{mur85a} that two algorithms are feasible.  
What is involved here is sketched out as follows.

Continuity-constrained single link hierarchical clustering is simultaneously 
hierarchical clustering on the spanning graph.  This is easy to implement
(inefficiently): just fix an infinite (or very large)  distance between 
non-contiguous pairs and proceed to use single link hierarchical clustering
\cite{mur85a}.

The complete link method, with the constraint that at least one member of 
each of the two clusters to be agglomerated be contiguous, is guaranteed 
not to give rise to inversions.   The $O(n^2)$ time, $O(n^2)$ space 
algorithm for the complete link method, based on the nearest neighbor chain
(see \cite{mur85a}), is easily modified to include an additional
testing of contiguity whenever a linkage in the nearest neighbor chain 
is created.

In this work we use the latter, in view of the well-balanced hierarchies
typically produced, \cite{mur84b}.

\section{Casablanca Film Script Analysis}
\label{casablanca}

\subsection{Data Used}

Based on the unpublished 1940 screenplay by Murray Burnett and Joan 
Allison, \cite{burnett}, the Casablanca script by Julius J. Epstein, 
Philip G. Epstein and Howard Koch led to the film directed by Michael 
Curtiz, produced by Hal B. Wallis and Jack L. Warner, and shot by 
Warner Bros.\ between May and August 1942.  We used the script from 
The Internet Movie Script Database, IMSDb (www.imsdb.com).  
 
The Casablanca film script comprises 77 successive scenes.  All told,
there were 6710 words in these successive scenes.  

The source text for the 77 scenes, including metadata, varied between
just 5 words, and 1017 words (scene 22).  
Typical Zipf distributional 
behavior is seen in the marginals.  
All words were used, following the imposing of lower 
case throughout.  This was subject only to words being longer than one 
character.  No stemming
or other preprocessing was used.  The top word frequencies were interesting: 
the, 965; rick, 
689; you, 651; to, 574; and, 435; in, 332; of, 319; renault, 284; it, 271; 
ilsa, 256; laszlo, 
255; he, 236; is, 232; at, 192; that, 171; for, 158; we, 151; on, 149; 
strasser, 135.  The numerically high presence
of personal names is quite unusual relative to more general texts, 
and characterizes this film script text.  A major reason for this is that 
character names head up each dialog block.  

Casablanca is based on a range of miniplots.  This occasions considerable
variety.  Miniplots include: love story, political drama, action 
sequences, urbane drama, and aspects of a musical.  The composition
of Casablanca 
is said by McKee \cite{mckee} to be ``virtually perfect'' (p. 287).  

\subsection{Analysis of Casablanca's ``Mid-Act Climax'', Scene 43}

To illustrate our methodology, applied below to scenes, we look first
in depth at one particular scene.

This scene is discussed in detail in McKee 
\cite{mckee}, who subdivides it into 
11 successive beats (here understood as subscenes).  It relates to 
Ilsa and Rick seeking black market exit visas.  Beat 1 is Rick finding 
Ilsa in the market.  Beats 2, 3 and 4 are rejections of him by Ilsa.  
Beats 5 and 6 express rapprochement by both characters.  Beat 7 is 
guilt-tripping by each in turn.  Beat 8 is a jump in content: Ilsa says
she will leave Casablanca soon.  In beat 9, Rick calls her a coward, 
and Ilsa calls him a fool.  Beat 10 is a major push by Rick, 
essentially propositioning her.  In the climax in beat 11, all goes to 
rack and ruin as Ilsa says she was married to Laszlo all along, and Rick 
counters that she is a whore.  

\begin{figure}
\includegraphics[width=14cm]{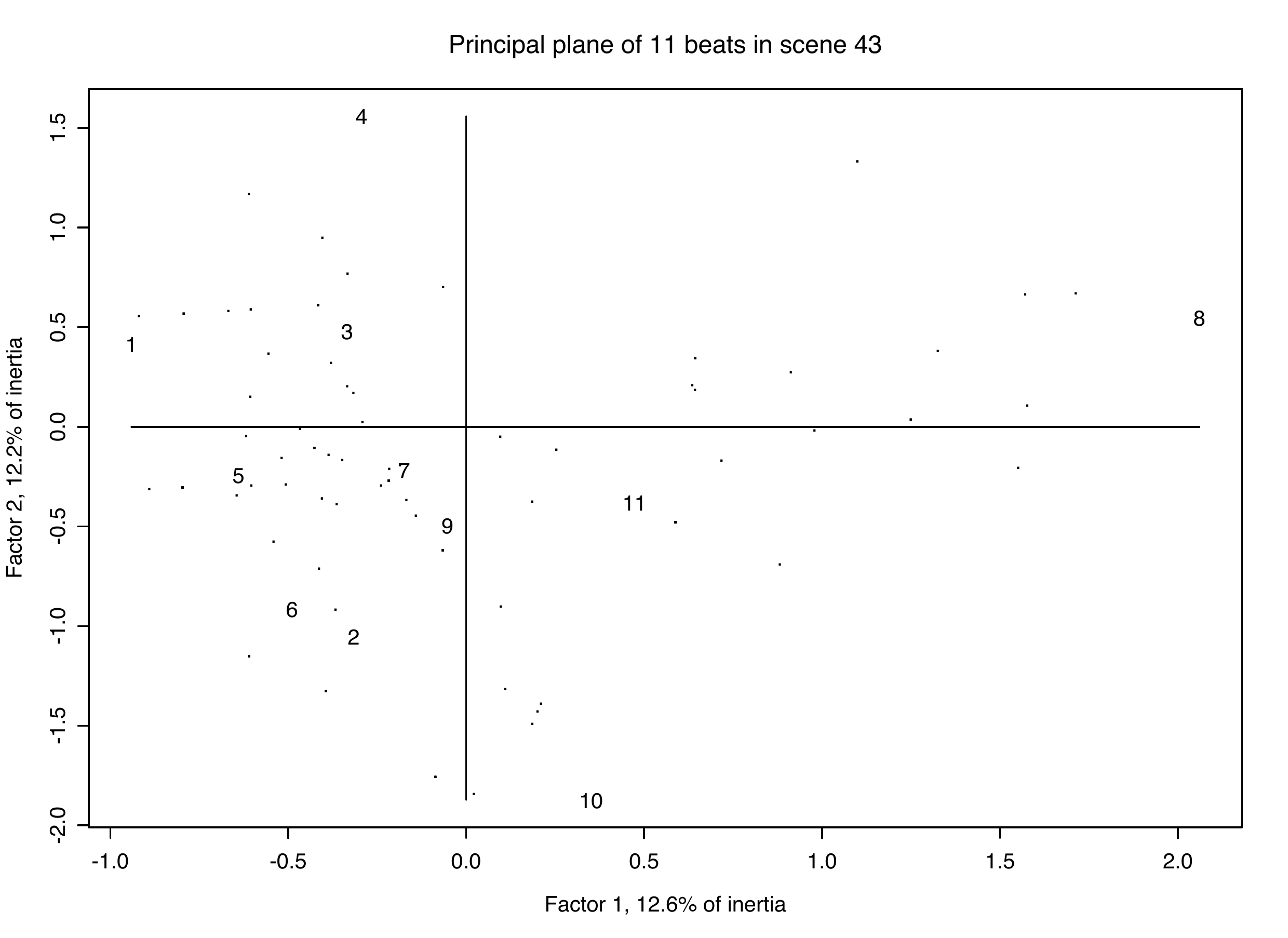}
\caption{Correspondence Analysis of the scene 43, which crosses (with 
presence/absence values) 11 successive beats (numbers) with, in total, 
210 words (dots: not labeled for clarity here).}
\label{fig1}
\end{figure}

\begin{figure}
\includegraphics*[width=14cm]{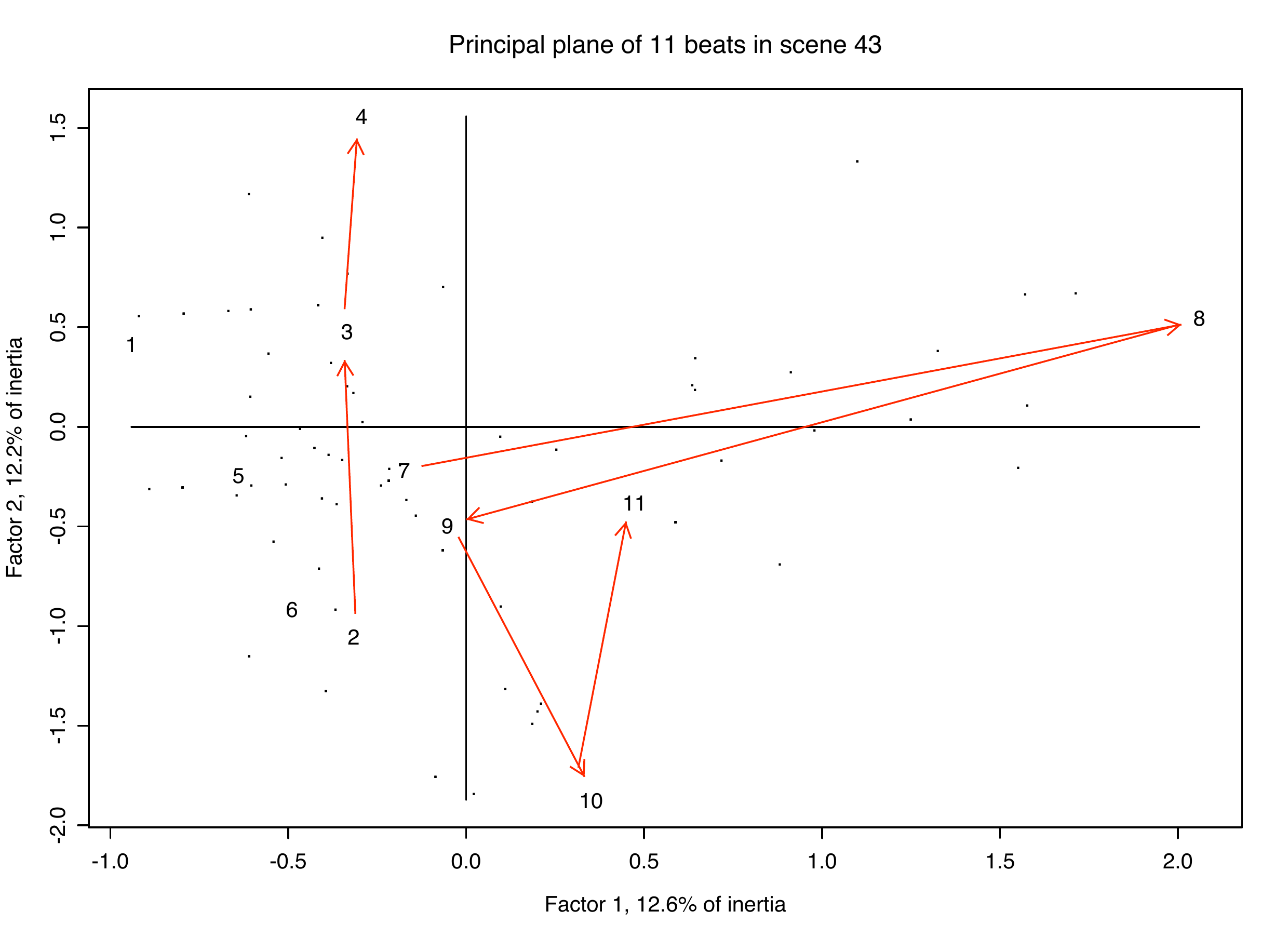}
\caption{As Figure \ref{fig1}, with some major movements from 
beat to beat noted and discussed in the text.}
\label{fig1annotated}
\end{figure}

Figure \ref{fig1} shows the best planar projection of the beats. The dots
show the locations of the words.  The importance of the factors is 
defined by the percentages of inertia explained by the factors (see 
Appendix for background details).  
We could certainly look further at 
what words are closest to scene 8.  (They are words having to do with 
Ilsa announcing that she will leave Casablanca.)   

Look though at the following aspects of Figure \ref{fig1}, 
as illustrated in Figure \ref{fig1annotated}.  Beats 2, 3 and 4 are moving 
nicely in one direction, so we can claim that moving in the positive 
direction of the ordinate (vertical axis) is  reinforcement of 
Ilsa's rejection of Rick.  As against this, movement in the negative
direction of the ordinate expresses rapprochement by Ilsa and 
Rick: look at beat 10.  Beat 8 is way off, and for Rick points to 
a real possibility of losing the game (of re-captivating Ilsa).  
The climax in beat 11 moves distinctly away from Rick's aspirations 
as expressed in beat 10.  

The length of the beat can show a lead-up to a climax in the scene, 
as noted in section \ref{delta}.  We see this very well in 
the beats of scene 43: the final five beats have lengths (in terms 
of presence of the words we use) of 50, 44, 38, 30, and then in
the climax beat, 46.  Earlier
beats vary in length, with successive 
word counts of 51, 23, 99, 39, 30, 17.  

The overall change in this scene, scene 43, 
is defined by the difference between
closing and opening beats.  Given the Correspondence Analysis output, 
where we have a Euclidean embedding taking care of weighting and 
normalization on both beat and word sets, we can easily take the 
full-dimensionality embedding (unlike the 2-dimensional projection 
seen in Figure \ref{fig1}) and determine this distance between 
beat number 1 and beat number 11.  As suggested very strongly 
by Figure \ref{fig1}, 
this distance will not necessarily be the greatest distance among 
successive beats.  

We reiterate that Figure \ref{fig1} provides us with a planar 
projection of the beats, which is optimal in a least squares 
sense, but is of necessity an approximation to the full-dimensionality 
clouds of beat, and word, points.  The quality of this best fit 
approximation is roughly 24.8\% (i.e., the sum of inertias explained
by the two axes of Figure \ref{fig1}) of the information content of 
the overall cloud, considered as either the cloud of beats, or the cloud
of words.  

For clustering the data displayed in Figure \ref{fig1} we will use 
the full dimensionality.  We have 
noted above some of the changes in direction in the succession of beats,
as displayed in Figure \ref{fig1}:
2, 3 and 4 following a particular sweep; 10 reverses this; and so on. 
Let us therefore look at the clustering of beats based solely on 
changes in direction or orientation.  In the full-dimensionality 
Correspondence Analysis embedding we will look not at the positions 
of the beats but instead at their correlation with the factors 
(i.e., axes or coordinates).  Changes from one beat's correlations 
with all factors, to those of the next beat, admirably express 
change in orientation of these successive beats.  

\begin{figure}
\includegraphics*[width=14cm]{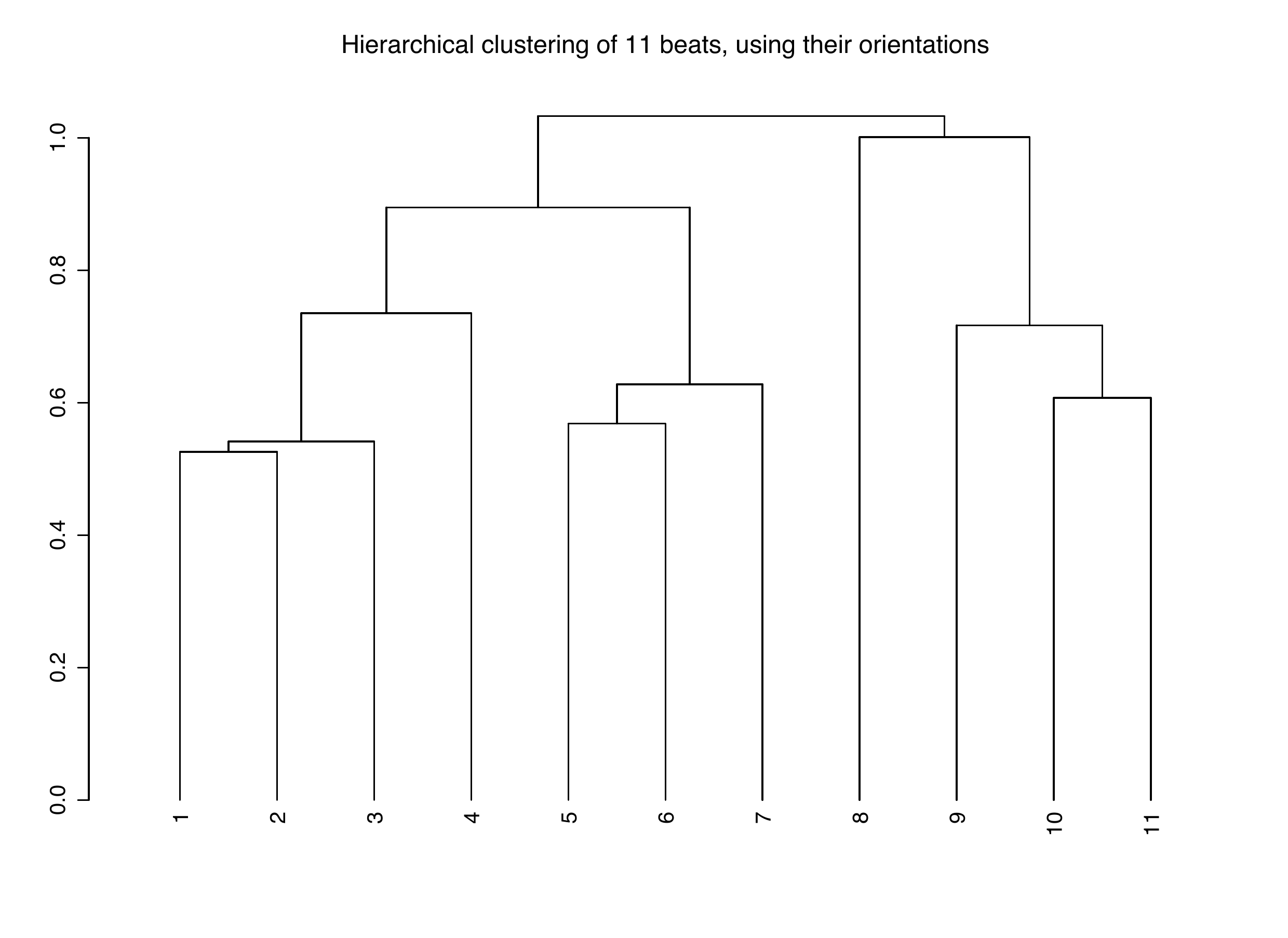}
\caption{Hierarchy of the 11 beats in scene 43, using the relationships
of beats defined from changes in orientation.}
\label{fig2}
\end{figure}

Figure \ref{fig2} shows the hierarchical clustering of the 
correlations (with all factors) for the 11 successive beats, 
using the sequence- or chronology-constrained agglomerative 
method discussed in section \ref{cclink}.  Note 
how beats 2, 3, 4 are clustered 
together; how 5, 6, 7 have a certain unity too; and in particular
how beat 8 is a sort of major caesura in the overall sequence of 
beats.  

We do not find everything needed to understand the beat succession 
of scene 43 of Casablanca in the vantage points offered by 
Figures \ref{fig1} and \ref{fig2}.  But we are gathering very 
useful perspectives on this scene.  To see how useful this is, 
let us carry out a benchmarking or baselining of what we 
see against an alternative of a randomized set of 11 beats in 
scene 43. 

\subsection{The Specific Style of a Film Script}
\label{style}

Arising out of our exploration so far, we will use the following
indicators of style and structure.  To be usable across different
film scripts, we must look at aggregate quantities.  Here we will
use first and second order moments.  We continue to use scene
43 of Casablanca, with its 11 successive, constituent beats.  

The attributes used are as follows.  

\begin{enumerate}
\item Attributes 1 and 2: The relative movement, given 
by the mean squared distance from one beat to the next.  We take
the mean and the variance of these relative movements.  Attributes 
1 and 2 are based on the (full dimensionality) factor space
embedding of the beats.  
\item Attributes 3 and 4: the changes in direction, given by the 
squared difference in correlation from one beat to the next.  
We take the mean and variance of these changes in direction.  
Attributes 3 and 4 are based on the (full dimensionality) 
correlations with factors.  
\item Attribute 5 is mean absolute tempo.  Tempo is given by 
difference in beat length from one beat to the next.  
Attribute 6 is the mean of the ups and downs of tempo.  
\item Attributes 7 and 8 are, respectively, the mean and variance
of rhythm given by the sums of squared deviations from one beat 
length to the next.  
\item Finally, attribute 9 is the mean of the rhythm taking up or
down into account.  
\end{enumerate}

For the Casablanca scene 43, we found the following as particularly
significant.  We tested the given scene, with its 11 beats, against
999 uniformly randomized sequences of 11 beats.  If we so wish, this 
provides  a
Monte Carlo significance test of a null hypothesis up to the 0.001 level.  

\begin{itemize}
\item In repeated runs, each of 999 randomizations, we find scene 43 to be
particularly significant (in 95\% of cases)
in terms of attribute 2: variability of movement from one 
beat to the next is smaller than randomized alternatives.  This may be
explained by the successive beats relating to coming together, or 
drawing apart, of Ilsa and Rick, as we have already noted.  
\item In 84\% of cases, scene 43 has greater tempo (attribute 5) 
than randomized
alternatives.  This attribute is related to absolute tempo, so we do 
not consider whether decreasing or increasing.  
\item In 83\% of cases, the mean rhythm (attribute 7) is higher than 
randomized alternatives.  
\end{itemize}

\subsection{Analysis of All 77 Scenes}

The clustering hierarchy that we focus on here is based on the 
orientation of the scenes.  This we do by taking any given scene's
correlations with the factors.  We have observed earlier that the
flow of the story, in relative terms, involves many ``backs and 
forths'' or ``tos and fros''.  This justifies our reason for looking at 
whether or not a group of scenes maintains an approximately 
similar orientation for some time, and how dramatic are the changes 
in direction.  

Our clustering algorithm takes the sequence of scenes into account.
As such, it offers a way to look at change over this sequence 
progression.  We could well construct such a hierarchy of changes 
on data other than scene orientation or direction.  We could use 
tempo or rhythm, for example.  However, orientation or direction 
serves us very well and provides, already, useful insight into 
deep structure.  

\begin{figure}
\includegraphics*[width=20cm,angle=270]{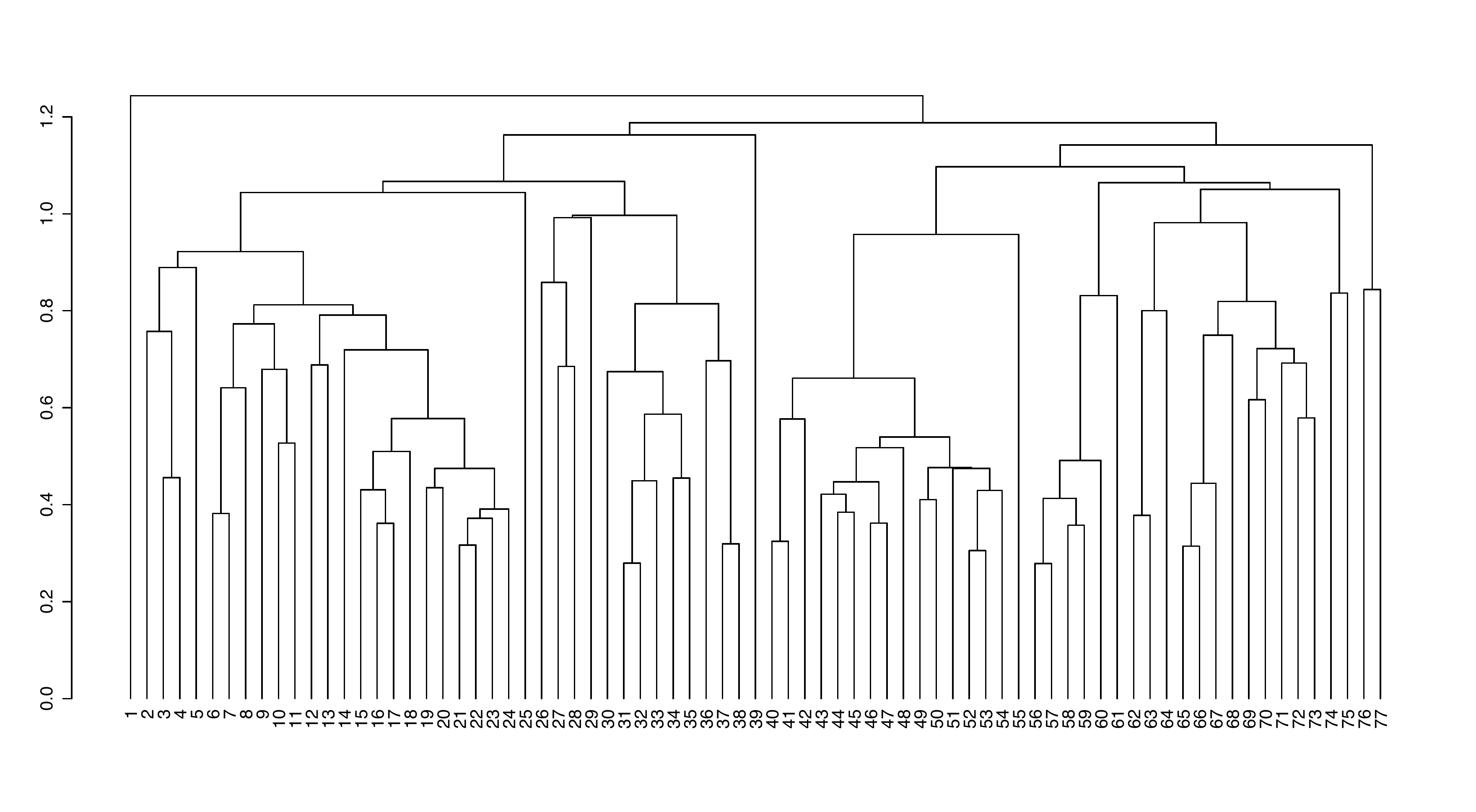}
\caption{Hierarchy of the 77 scenes, respecting the sequence, based on the 
characterization of the 77 scenes by the set of all words used in the scenes.
The directional information (i.e., correlations with the factors)
associated with the scenes is used.}
\label{fig6}
\end{figure}

In Figure \ref{fig6} we see how different scene 1 is, relating to 
narrated scene-setting of the Second World War. Scene 25 is a flashback
to Paris in the spring.  Scene 39 is set in a black market in Casablanca, 
where a Native and a Frenchman appear, but not any of the central 
characters.  In this hierarchy we can see a pronounced redirection 
of the story in scenes 38, 39 and 40.  (Note how the ultrametric
distance between scene 39, on the one hand, and scene 38 or any 
preceding scene on the other hand, is relativley very great.  The 
ultrametric distance between scene 39 and scene 40 is even greater.)
In scene 38, Laszlo and Ilsa 
are in Renault's office, clarifying their visa situation.   The black 
market of scene 39 points the finger at Signor Ferrari, to be 
found at the Blue Parrot cafe.  Scene 40 is then at the Blue Parrot.  
The essential issue of Laszlo's role and problem of getting a visa is 
revealed in these scenes.  
In the overall story-line, these scenes are more or less right at the 
mid-point.  The pairing off of scenes at the end (74 and 75, 76 and 
77: respectively, airport, hangar, road, hangar) is very much in keeping 
with the content of these scenes.  
Together they represent the climax scenes.

A potential use of Figure \ref{fig6} is to provide an indication of 
possible commercial breaks between acts or sequences of scenes, such 
that these breaks are derived automatically from the screenplay and without 
the writer explicitly marking them in the text.  In cinema movie 
such breaks are not pre-planned.  The hierarchy provides
a visualization allowing comparison between the writer's intentions and
one (albeit insightful)  view of where these breaks are found to be
located, coupled with the strength of the breaks.  

We again looked as style and structure, using 999 randomizations of 
the sequence of 77 scenes.  Some interesting conclusions were garnered.

\begin{itemize}
\item As for the case of beats in scene 43, we find that the entire 
Casablanca plot is well-characterized by the variability of movement 
from one scene to the next (attribute 2).  Variability of movement from one
beat to the next is smaller than randomized alternatives in 82\% of cases.
\item Similarity of orientation from one scene to the next 
(attribute 3) is very tight,
i.e.\ smaller than randomized alternatives.  We found this to hold in 
95\% of cases.  The variability of orientations (attribute 4)
was also tighter, in 82\% of cases. 
\item Attribute 6, the mean of ups and downs of tempos is also revealing.
In 96\% of cases, it was smaller in the real Casablanca, as opposed to 
the randomized alternatives.  This points to the ``balance'' of up 
and down movement in pace.  
\end{itemize}

\section{Television Series Script Analysis}
\label{csi}

Our discussion so far has been for the Casablanca movie.  Now we
turn to television drama, for which other constraining aspects hold
(such as length, and inter- as well as intra-cohesion and homogeneity).  

We took three CSI (Crime Scene Investigation, Las Vegas -- Grissom, Sara, 
Catherine  et al.) television scripts from series 1:

\begin{itemize} 
\item 1X01, Pilot, original air date
on CBS Oct. 6, 2000.  Written by Anthony E. Zuiker, directed by Danny
Cannon.   
\item 1X02, Cool Change, original air date on CBS, Oct. 
13, 2000.  Written by Anthony E. Zuiker, directed by Michael Watkins.
\item 1X03, Crate 'N Burial, 
original air date on CBS, Oct. 20, 2000.  Written by Ann Donahue, directed
by Danny Cannon.  
\end{itemize}

Note the  differences between
writers and directors in most cases.  This lends weight to our 
goal of furnishing the producer and director teams with a platform
for automatically or semi-automatically assessing quality of 
product.
We will refer to these scripts as 
CSI-101, CSI-102 and CSI-103.   All film scripts were obtained 
from TWIZ TV (Free TV Scripts \& Movie Screenplays Archives), 
http://twiztv.com

From series 3, we took another three scripts. 

\begin{itemize}
\item 3X21, Forever, original air
date on CBS, May 1, 2003.  Written by Sara Goldfinger, directed by David 
Grossman.
\item 3X22, Play With Fire, original air date on CBS,
May 8, 2003.  Written by Naren Shankar and Andrew Lipsitz, directed by 
Kenneth Fink.
\item 3X23, Inside The Box, original air date on CBS, May 15, 
2003.  Written by Carol Mendelsohn and Anthony E. Zuiker, directed by Danny 
Cannon.
\end{itemize}

We will refer to these as CSI-321, CSI-322 and CSI-323.

An example of a very short scene, scene 25 from CSI-101, follows.  

\begin{footnotesize}
\begin{verbatim}

[INT. CSI - EVIDENCE ROOM -- NIGHT]

(WARRICK opens the evidence package and takes out the shoe.)

(He sits down and examines the shoe.  After several dissolves, WARRICK 
opens the
lip of the shoe and looks inside.  He finds something.)

WARRICK BROWN:  Well, I'll be damned.

(He tips the shoe over and a piece of toe nail falls out onto the table.  He
picks it up.)

WARRICK BROWN:  Tripped over a rattle, my ass.

\end{verbatim}
\end{footnotesize}

We see here scene metadata, characters, dialog, and action information,
all of which we use. 
Frontpiece, preliminary or preceding storyline information,
and credits were ignored by us.  We took the labeled scenes.  
The number of scenes in each movie, and the number of 
unique, 2-characters or more, words used in the movie, are listed in 
Table \ref{tab0}.  
All punctuation was ignored.  All upper case was 
converted to lower case.  Otherwise there was no pruning of stopwords.  
The top words and their frequencies of occurrence were:

\bigskip 

\noindent the 443; to 239; grissom 195; you 176; and 166; gil 114;
catherine 105; of 89; he 85; nick 80; in 79; on 79; it 78; at 76;
ted 66; sara 65; warrick 65; ...

\bigskip

\begin{table}
\begin{center}
\begin{tabular}{ccc}  \hline
Script     &  No.\ scenes  &   No.\ words  \\ \hline
CSI-101    &     50        &    1679        \\
CSI-102    &     37        &    1343        \\
CSI-103    &     38        &    1413        \\  
CSI-321    &     39        &    1584        \\
CSI-322    &     40        &    1579        \\
CSI-323    &     49        &    1445        \\ \hline
\end{tabular}
\end{center}
\caption{Numbers of scenes in the plot, and numbers of unique (2-letter
or more) words.}
\label{tab0}
\end{table}

In order to equalize the Zipf distribution of words, and to homogenize 
the scenes and words by considering {\em profiles} (as opposed to 
raw data), as before we 
embedded the set of scenes in a Euclidean factor space in all cases,
using Correspondence Analysis.  We then clustered the full dimensionality 
factor space,
using the hierarchical agglomerative algorithm that took into account the 
sequence of scenes, viz.\ the sequence-constrained
complete link agglomerative method.  
To capture the ``drift'' of direction of the story, again like before, 
we used correlations rather than projections.  

\begin{figure}
\begin{center}
\includegraphics*[width=18cm,angle=270]{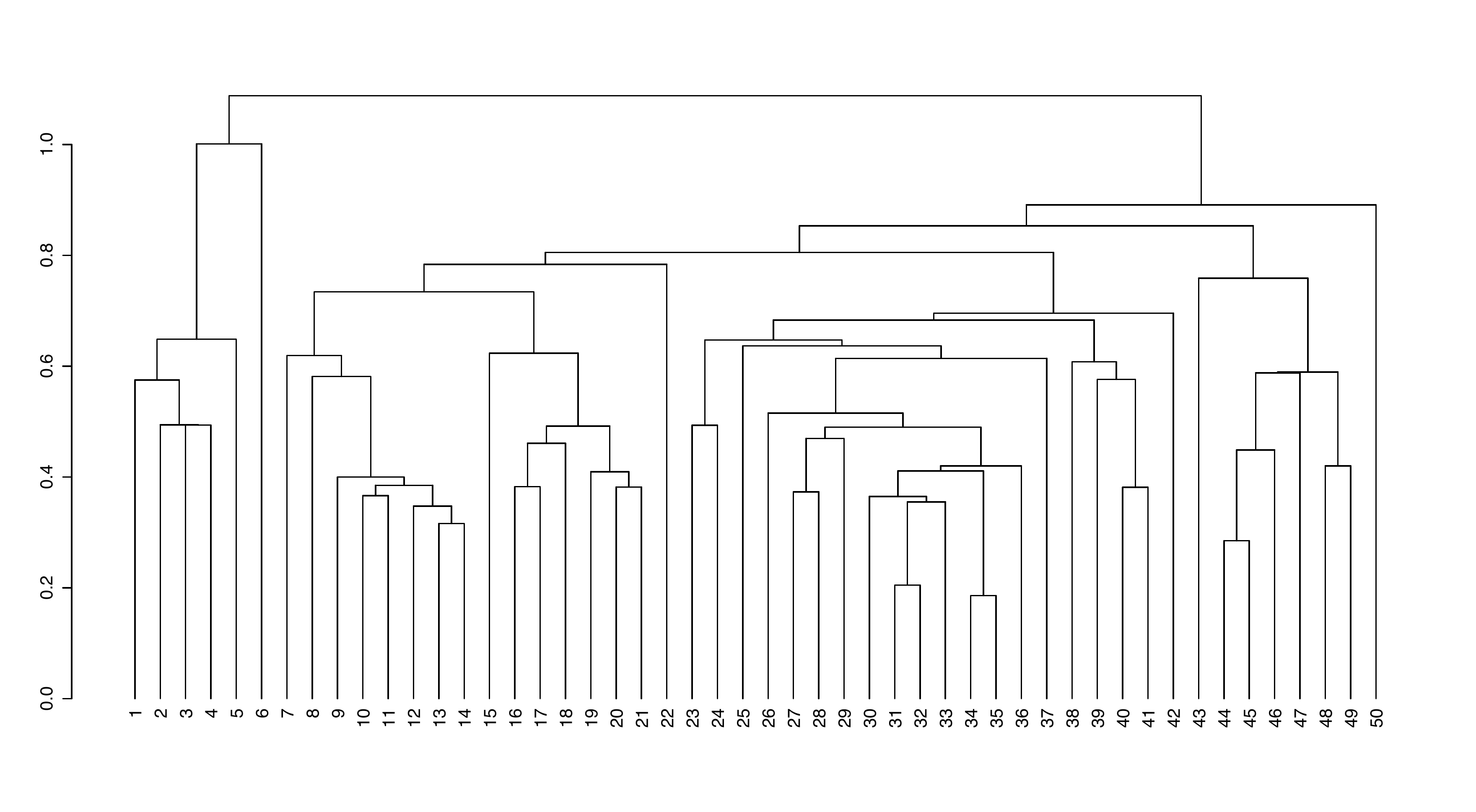}
\end{center}
\caption{Sequence-constrained complete link hierarchy for CSI episode 101.
Orientations (full dimensionality) were used for each of the 50 scenes.}
\label{fig82}
\end{figure}

\begin{figure}
\begin{center}
\includegraphics*[width=18cm,angle=270]{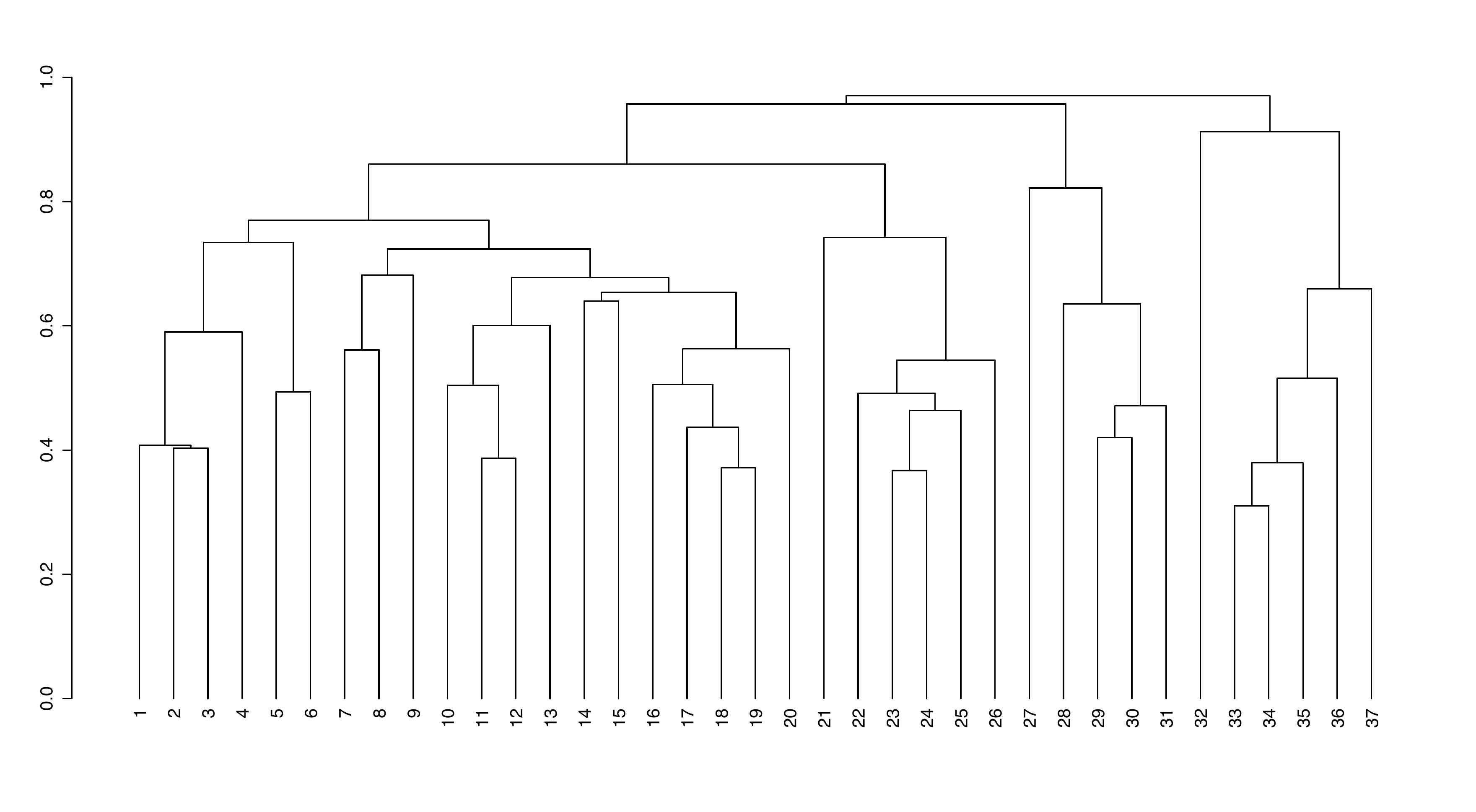}
\end{center}
\caption{Sequence-constrained complete link hierarchy for CSI episode 102.
Orientations (full dimensionality) were used for each of the 37 scenes.}
\label{fig84}

\end{figure}
\begin{figure}
\begin{center}
\includegraphics*[width=18cm,angle=270]{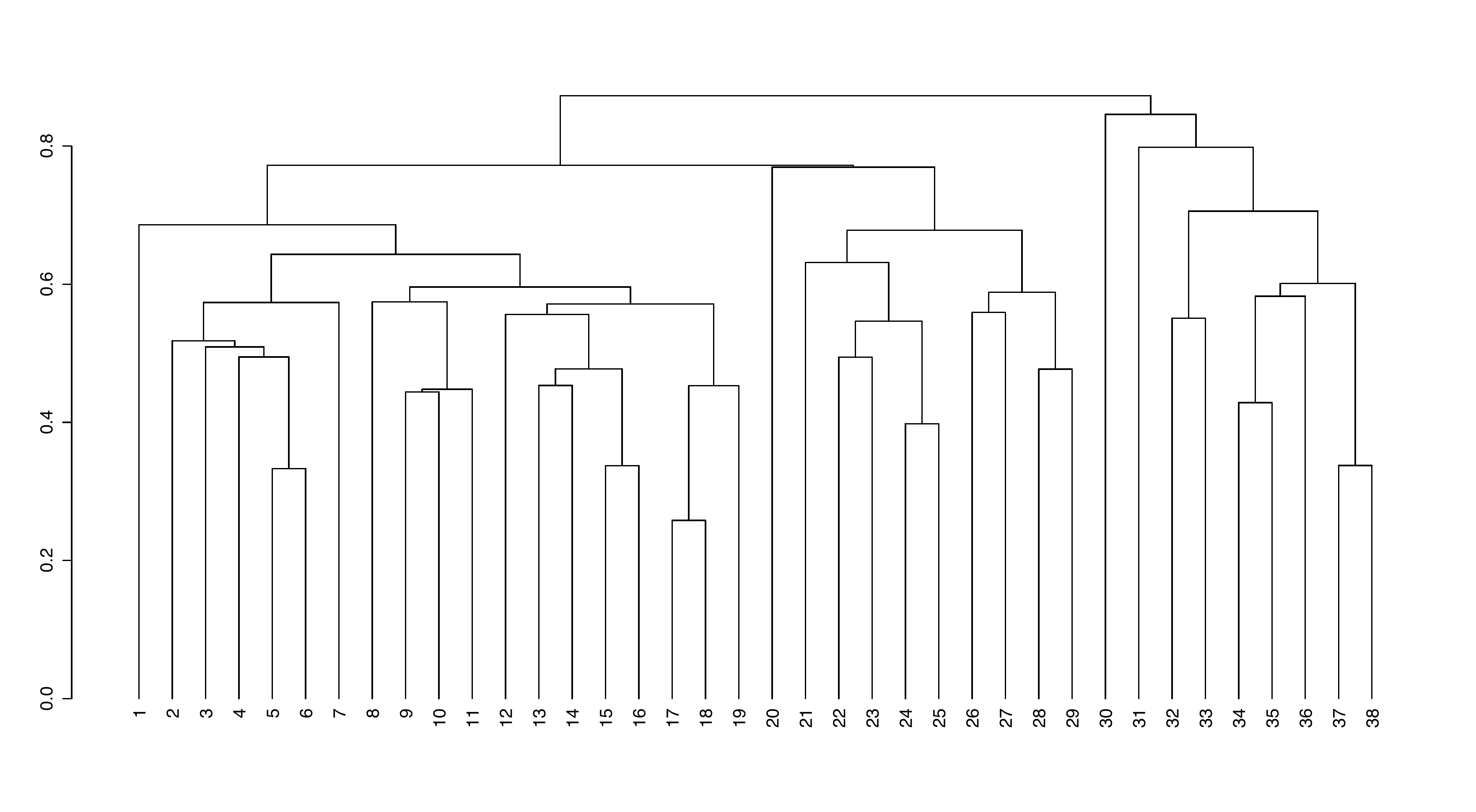}
\end{center}
\caption{Sequence-constrained complete link hierarchy for CSI episode 103.
Orientations (full dimensionality) were used for each of the 38 scenes.}
\label{fig86}
\end{figure}

Figures \ref{fig82}, \ref{fig84} and \ref{fig86} show the results 
obtained.   

To focus discussion of internal structure which can be appreciated
in these figures, let us look at where commercial breaks are flagged. 
While other considerations are important, like elapsed time from the 
start, it is clear that continuity of content is also highly relevant.
Television episodes are written to create minor cliffhangers at the 
commercial breaks and therefore if the breaks can be identified by 
our datamining approach there is prima facie evidence for the finding
of deep structures within screenplays.   
Four commercial breaks are flagged in the 
script of CSI-101, and three in the scripts of CSI-102 and CSI-103.  

For CSI-101, Figure \ref{fig82}, 
these commercial breaks, as given in the script, were between scenes 
4 and 5; 14 and 15 (substantial change noticeable in Figure \ref{fig82}); and
32 and 33.   The change in direction in the climax scene, scene 50, is 
clear.  The early scenes, 1 up to 6, are distinguishable from the scenes
that follow.  

For CSI-102, Figure \ref{fig84}, 
the commercial breaks were between scenes 4 and 5; 12 and 13; and 22 
and 23.  The climax here appears to be a collection of scenes, from 
33 to 37.

Finally, for CSI-103, Figure \ref{fig86}, 
the commercial breaks were between scenes 1 and 2 (clear change); 
9 and 10; and 29 and 30 (clear change).  

We summarize these findings as follows.  There is occasionally a 
very strong link between commercial breaks and change in 
thematic content as evidenced by the hierarchy.  In other cases
we find continuity of content bridging the gap of the commercial 
breaks.  

Programs CSI-102 and CSI-103 are perhaps clearer than CSI-101 in 
having a more balanced subdivision.  For CSI-102, we would put this 
subdivision as 
from scenes 1 to 20; 21 to 26; 27 to 31; and 32 to 37.  For CSI-103,
we would demarcate the plot into scenes 1 to 19; (20 or) 21 to 29; and
30 to 38.

As in section \ref{style}, we looked at the characteristics of style in 
the scripts.  For a given script, we characterized it on the basis of our
nine attributes.  Then we randomized the order of the scenes comprising 
the script.  So the plot (or story) was identical in terms of the scenes that
constitute it.  But the plot was lacking in sense -- in style and 
in structure -- to the extent that the scenes were now in a random order.
Such a randomized plot was also characterized on the basis of our nine
attributes.  We carried out 999 such randomizations.  When an attribute's
value for the real script was found to be less than or greater than 
80\% of the randomized plots, then we report it in Table \ref{tabtab}.
Our significance threshold of 80\% was set at this value to be 
sufficiently decisive.  It was rounded to an integer percentage in all
cases.  

\begin{table}
\begin{center}
\begin{tabular}{ccccc} \hline 
program script & attribute  &  program $\leq / \geq$ &  \% of cases  \\ \hline
               &           &                          &             \\
CSI-101        &   1       &    $\leq$                &  87\%        \\
CSI-101        &   3       &    $\leq$                &  93\%        \\
CSI-101        &   5       &    $\leq$                &  84\&        \\
CSI-101        &   6       &    $\geq$                &  84\%        \\
CSI-101        &   7       &    $\leq$                & 90\%        \\
CSI-101        &   9       &    $\leq$                &  88\%       \\
               &           &                          &             \\
CSI-102        &   1       &    $\leq$                &  95\%       \\
CSI-102        &   2       &    $\leq$                &  95\%       \\
CSI-102        &   3       &    $\leq$                &  95\%       \\
CSI-102        &   5       &    $\geq$                &  81\%       \\
CSI-102        &   6       &    $\leq$                &  81\%       \\
               &           &                          &             \\
CSI-103        &   2       &    $\leq$                &  88\%       \\
CSI-103        &   3       &    $\leq$                &  95\%       \\
CSI-103       &    4      &    $\geq$                &  83\%       \\
CSI-103        &   6       &    $\leq$                &  88\%       \\
CSI-103        &   9       &    $\leq$                &  88\%       \\
               &           &                          &             \\
CSI-321        &   1       &    $\leq$                &  83\%       \\
CSI-321        &   2       &    $\geq$                &  91\%       \\
               &           &                          &             \\
CSI-322        &   1       &    $\leq$                &  92\%       \\
CSI-322        &   3       &    $\leq$                &  97\%       \\
CSI-322        &   4       &    $\leq$                &  86\%       \\
CSI-322        &   6       &    $\geq$                &  86\%       \\
               &           &                          &             \\
CSI-323        &   1       &    $\leq$                &  92\%       \\
CSI-323        &   8       &    $\leq$                &  81\%       \\
CSI-323        &   9       &    $\leq$                &  81\%       \\ \hline
\end{tabular}
\end{center}
\caption{Shown are how and where the six television movies considered
are unique in structure and style.  
For a particular attribute (see section \ref{style} for description)
the program script was different from randomized scene sequences 
in the indicated \% of cases.}
\label{tabtab}
\end{table}

In regard to Table \ref{tabtab} we recall from section \ref{style} that
attributes 1 and 2 are first and second moments, respectively, of 
relative movement from one scene to the next.  Attributes 3 and 4 are 
first and second moments of relative orientation from one scene to the next.
Attributes 5 and 6 relate to tempo.  Attributes 7, 8 and 9 relate to rhythm.  

Furthermore whether our script is less than or greater than the 
randomized alternatives -- cf.\ column 3 of Table \ref{tabtab} -- can 
be understood as follows.  If the ``less than or equal to'' case applies
we can view this as our script being more compact or more 
parsimonious or more smooth or low frequency, for the particular 
attribute at issue, 
relative to the great bulk of randomized alternatives.  Where the 
``greater than or equal to'' case applies, then we can see 
something exceptional in the way that the plot is handled.  

In Table \ref{tabtab}, attribute 1 (mean relative movement) is 
a strong characterizing marker for all scripts, save one.  This attribute
is ``compact'' for the real script (in the sense in which we have used
this term of ``compact'' in the last paragraph, with reference to 
column 3 of Table \ref{tabtab}).   Attribute 3 (mean relative 
reorientation) is a good characterizing marker for four of the 
six scripts.  Attribute 9 (rhythm) is also a good marker for three of 
the six scripts. 

Our Monte Carlo procedure is a rigorous one for assessing significance 
of patterns in the filmscript data.  As we have demonstrated it 
allows us to validate unique semantic properties underlying the
``sensory surface'' (McKee) of the filmscripts.

\section{Conclusions}

The basis for accessing semantics in provided by (i) Correspondence 
Analysis, where each scene is an average of words or other attributes
that characterize it, and each attribute is an average of scenes that 
are characterized; and (ii) in the hierarchical clustering of the 
sequence of scenes, relative change is modeled by the dendrogram structure.  

We have made excellent progress in this work on having the 
qualitative precepts of 
McKee \cite{mckee} both quantified and operationalized.  Our 
assessments of the Casablanca movie, and the six CSI episodes, 
show that there is a great deal of commonality in style and 
structure  between film and television.  

We have taken into account both the linear and the hierarchical
relationships in the plot, expressing the story.  The units used 
were beats (i.e., subscenes) 
and scenes, essentially, with the hierarchical clusterings
revealing larger scale structures (beginning scenes; climax scenes; 
and the halves, or thirds, or whatever segments were revealed as 
appropriate for the entire plot).  

Let us look now at how our work is of importance for the study of 
style and structure in narrative, in general.  

 Chafe \cite{chafe}, in analyzing verbalized memory,  used
a 7-minute 16 mm color movie, with sound but no language, and collected
narrative reminiscences of it from human subjects, 60 of whom were
English-speaking and at 
least 20 spoke/wrote one of nine other languages.  Chafe considered 
the following units.

\begin{enumerate}
\item {\em Memory} expressed by a {\em story} (memory takes the form of an
``island''; it is  ``highly selective''; it is a ``disjointed chunk''; but
it is not a book, nor a chapter, nor a continuous record, nor a stream).
\item {\em Episode}, expressed by a {\em paragraph}.
\item {\em Thought}, expressed by a {\em sentence}.
\item A {\em focus}, expressed by a {\em phrase} (often these 
phrases are linguistic ``clauses'').
Foci are ``in a sense, the basic units of memory in that they represent
the amount of information to which a person can devote his central
attention at any one time''. 
\end{enumerate}

The ``flow of thought and the flow of language'' are treated at once,
the latter proxying the former, and analyzed in their linear and
hierarchical structure by \cite{chafe,hinds,longacre}, 
among others.  
Filmscript affords us
clear boundaries between the units of text that are analyzed.
For more general text, we must consider segmentation.  
Examples of text 
segmentation to open up the analysis of style and structure include
\cite{hearst,bestgen,groszsidner,grosz2002,choi,skorochodko}.  

We have shown in this work how useful the story 
expressed in a film or television movie script can be, in order to 
provide a framework for analysis of style and structure.  

\section*{Appendix: Correspondence Analysis and Hierarchical Clustering}

\subsection*{Analysis Chain}

Correspondence Analysis, in conjunction with hierarchical clustering, 
provides what could be characterized as a data analysis platform
providing access to the semantics of information expressed by the data.  
The way it does this is (i) by viewing each observation or row vector as the 
average of all attributes that are related to it; and by viewing each
attribute or column vector as the average of all observations that are 
related to it; and (ii) by taking into account the 
clustering and dominance relationships given by the hierarchical clustering.   

The analysis chain is as follows:

\begin{enumerate}
\item The starting point is a matrix that cross-tabulates the dependencies,
e.g.\ frequencies of joint occurrence, of an observations crossed by attributes
matrix.  
\item By endowing the cross-tabulation matrix with the $\chi^2$ metric 
on both observation set (rows) and attribute set (columns), we can map 
observations and attributes into the same space, endowed with the Euclidean
metric.  
\item A hierarchical clustering is induced on the Euclidean space, the 
factor space.  
\item Interpretation is through projections of observations, attributes 
or clusters onto factors.  The factors are ordered by decreasing importance.
\end{enumerate}

There are various aspects of Correspondence Analysis which follow on 
from this, such as Multiple Correspondence Analysis, different ways that
one can encode input data, and mutual description of clusters in terms of 
factors and vice versa.  See \cite{ref08888} and references therein 
for further details.  

We will use a very succinct and powerful tensor notation in the following, 
introduced by \cite{benz}.  At key points we will indicate the 
equivalent vector and matrix expressions.  

\subsection*{Correspondence Analysis: 
Mapping $\chi^2$ Distances into Euclidean Distances}

The given contingency table (or numbers of occurrence) 
data is denoted $k_{IJ} =
\{ k_{IJ}(i,j) = k(i, j) ; i \in I, j \in J \}$.  $I$ is the set of 
observation 
indexes, and $J$ is the set of attribute indexes.  We have
$k(i) = \sum_{j \in J} k(i, j)$.  Analogously $k(j)$ is defined,
and $k = \sum_{i \in I, j \in J} k(i,j)$.  Next, $f_{IJ} = \{ f_{ij}
= k(i,j)/k ; i \in I, j \in J\} \subset \R_{I \times J}$,
similarly $f_I$ is defined as  $\{f_i = k(i)/k ; i \in I, j \in J\}
\subset \R_I$, and $f_J$ analogously.  What we have described here is 
taking numbers of occurrences into relative frequencies.

The conditional distribution of $f_J$ knowing $i \in I$, also termed
the $j$th profile with coordinates indexed by the elements of $I$, is:

$$ f^i_J = \{ f^i_j = f_{ij}/f_i = (k_{ij}/k)/(k_i/k) ; f_i > 0 ;
j \in J \}$$ and likewise for $f^j_I$.  

\subsection*{Input: Cloud of Points Endowed with the Chi Squared Metric}

The cloud of points consists of the couples: 
(multidimensional) profile coordinate and (scalar) mass.
We have $ N_J(I) = \{ ( f^i_J, f_i ) ; i  \in I \} \subset \R_J $, and
again similarly for $N_I(J)$.  Included in this expression is the fact
that the cloud of observations, $ N_J(I)$, is a subset of the real 
space of dimensionality $| J |$ where $| . |$ denotes cardinality 
of the attribute set, $J$.  

The overall inertia is as follows: 
$$M^2(N_J(I)) = M^2(N_I(J)) = \| f_{IJ} - f_I f_J \|^2_{f_I f_J} $$
\begin{equation}
= \sum_{i \in I, j \in J} (f_{ij} - f_i f_j)^2 / f_i f_j
\label{eqnin}
\end{equation}
The term  $\| f_{IJ} - f_I f_J \|^2_{f_I f_J}$ is the $\chi^2$ metric
between the probability distribution $f_{IJ}$ and the product of marginal
distributions $f_I f_J$, with as center of the metric the product
$f_I f_J$.  Decomposing the moment of inertia of the cloud $N_J(I)$ -- or 
of $N_I(J)$ since both analyses are inherently related -- furnishes the 
principal axes of inertia, defined from a singular value decomposition.

\subsection*{Output: Cloud of Points Endowed with the Euclidean 
Metric in Factor Space}

The $\chi^2$ distance with center $f_J$ between observations $i$ and 
$i'$ is written as follows in two different notations: 

\begin{equation}
d(i,i') = \| f^i_J - f^{i'}_J \|^2_{f_J} = \sum_j \frac{1}{f_j} 
\left( \frac{f_{ij}}{f_i} - \frac{f_{i'j}}{f_{i'}} \right)^2
\end{equation}

In the factor space this pairwise distance is identical.  The coordinate
system and the metric change.  For factors indexed by $\alpha$ and for 
total dimensionality $N$ ($ N = \mbox{ min } \{ |I| - 1, |J| - 1 \}$;
the subtraction of 1 is since 
the $\chi^2$ distance is centered and  
hence there is a linear dependency which 
reduces the inherent dimensionality by 1) we have the projection of 
observation $i$ on the $\alpha$th factor, $F_\alpha$, given by 
$F_\alpha(i)$: 

\begin{equation}
d(i,i') = \sum_{\alpha = 1..N} \left( F_\alpha(i) - F_\alpha(i') \right)^2
\end{equation}

In Correspondence Analysis the factors are ordered by decreasing 
moments of inertia.  The factors are closely related, mathematically, 
in the decomposition of the overall cloud, 
$N_J(I)$ and $N_I(J)$, inertias.  The eigenvalues associated with the 
factors, identically in the space of observations indexed by set $I$, 
and in the space of attributes indexed by set $J$, are given by the 
eigenvalues associated with the decomposition of the inertia.  The 
decomposition of the inertia is a 
principal axis decomposition, which is arrived at through a singular
value decomposition.  

\subsection*{Hierarchical Clustering}

Background on the theory and practice of hierarchical clustering
can be found in \cite{mur85a,ref08888}. 
For the particular hierarchical clustering algorithm used here,
based on the given sequence of observations, \cite{mur85a} should 
be referred to.  A short description follows.  

Consider the projection of observation $i$ onto the set of 
all factors indexed by $\alpha$, $\{ F_\alpha(i) \}$ for all $\alpha$,
which defines the observation $i$ in the new coordinate frame.  
This new factor space is 
endowed with the (unweighted) Euclidean distance,  $d$.  
We seek a hierarchical clustering that takes into account the 
observation  sequence,
i.e.\ observation $i$ precedes observation $i'$ for all $i, i' \in I$.  
We use the linear
order on the observation.  Let us switch to the term {\em texts} now, 
which is what our observations refer to in this work.  
We refer to ``adjacent'' texts when one follows the 
other with respect to this linear order, and this definition of adjacency 
is extended to allow for adjacent clusters of texts.   

The agglomerative hierarchical clustering algorithm is as follows.  

\begin{enumerate}
\item Consider each text in the sequence of texts as constituting a 
singleton cluster.  
Determine the closest pair of adjacent texts, and define a cluster
from them.
\item Determine and merge 
the closest pair of adjacent clusters, $c_1$ and $c_2$, 
where closeness is defined by $d(c_1, c_2) = \mbox{ max } \{ d_{ii'}
\mbox{ such that } i \in c_1, i' \in c_2 \}$.  
\item Repeat step 2 until only one cluster remains.
\end{enumerate}

Here we use  a complete link criterion which additionally takes account of
the adjacency constraint imposed by the sequence of texts in set $I$.  
It can be shown (see \cite{mur85a}) 
that the closeness value, given by $d$, at each 
agglomerative step is strictly non-decreasing.  That is, if cluster $c_3$ 
is formed earlier in the series of agglomerations compared to cluster 
$c_4$, then the corresponding distances will satisfy $d_{c3} \leq 
d_{c4}$.  ($d$ here is as determined in step 2 of the algorithm above.)

\end{document}